\documentclass{article}
\usepackage{spconf,amsmath,graphicx}

\usepackage{amsfonts}
\usepackage{subfigure}

\title{Adaptive Parameter Sharing for Multi-agent Reinforcement Learning}
\name{Dapeng Li, Na Lou, Bin Zhang, Zhiwei Xu, Guoliang Fan}
\address{The Key Laboratory of Cognition and Decision Intelligence for Complex Systems, \\ Institute of Automation, Chinese Academy of Sciences\\
School of Artificial Intelligence, University of Chinese Academy of Sciences}
\begin{document}
%
\maketitle
\begin{abstract}
Parameter sharing, as an important technique in multi-agent systems, can effectively solve the scalability issue in large-scale agent problems. However, the effectiveness of parameter sharing largely depends on the environment setting. When agents have different identities or tasks, naive parameter sharing makes it difficult to generate sufficiently differentiated strategies for agents. Inspired by research pertaining to the brain in biology, we propose a novel parameter sharing method. It maps each type of agent to different regions within a shared network based on their identity, resulting in distinct subnetworks. Therefore, our method can increase the diversity of strategies among different agents without introducing additional training parameters. Through experiments conducted in multiple environments, our method has shown better performance than other parameter sharing methods.
\end{abstract}
\section{Introduction}
\label{sec:intro}
\vspace{-0.2cm}

Multi-agent reinforcement learning (MARL) aims to learn strategies that enable multiple agents to solve specific tasks in complex environments. In recent years, MARL has demonstrated notable success across various domains, including traffic signal control~\cite{trafic_chu}, game decision-making~\cite{xu2023dual,starcraft2}, and communication transmission~\cite{Busoniu2010}. While MARL has achieved success in settings with a small number and single type of agents, the research on addressing large-scale multi-agent problems involving multiple types is still ongoing~\cite{li2023sea}. Adapting existing MARL algorithms to handle a larger number of agents is often challenging, mainly due to the parameter growth with the increase in the number of agents.

Parameter sharing~\cite{gupta2017cooperative,li2023style} can help address the scalability issue faced by large-scale multi-agent systems. Naive parameter sharing simply shares the policy network parameters among all agents, enhancing sample efficiency and alleviating the problem of parameter growth as the number of agents increases. However, naive parameter sharing assumes that all agents have similar characteristics and reward functions. To address the problem in heterogeneity of multi-agent systems, a more effective way is selective parameter sharing ~\cite{seps}, in which agents under the same type share one independent network. Unfortunately, this technique does not address the fundamental issue of parameter growth in large-scale multi-agent systems. The size of the training parameters still increases with the number of agent types. Hence, it becomes imperative to devise an approach that not only guarantees distinct expressive capabilities for various agent types but also effectively handles the dimensionality of training parameters.

\begin{figure}[t]
  \centering
  \centerline{\includegraphics[width=8cm]{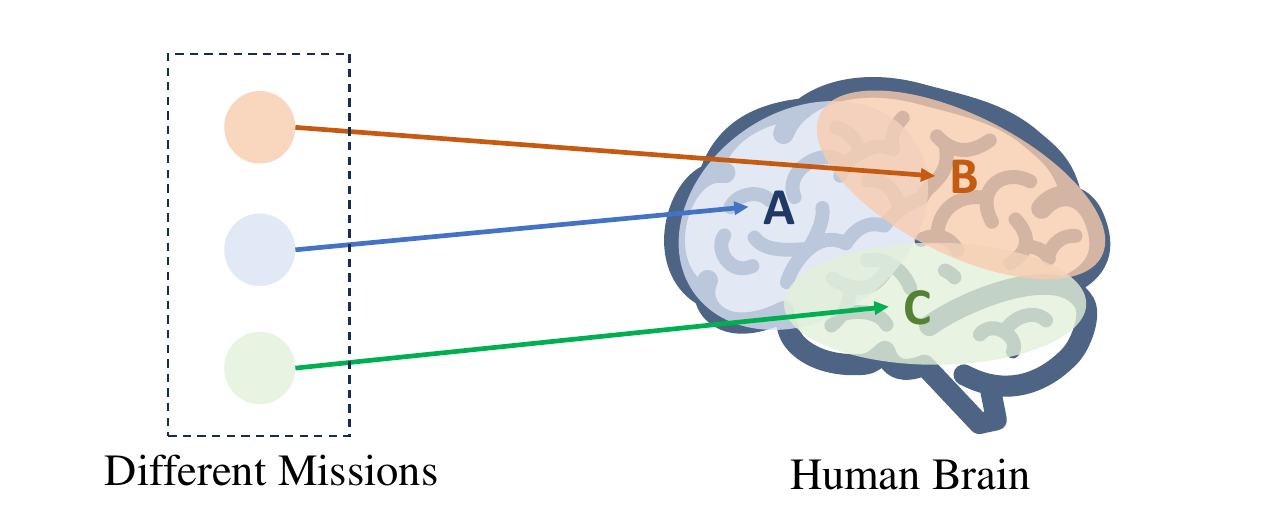}}
  \vspace{-0.3cm}
  \caption{Tasks mapping to specific region in human brains.}
  \vspace{-0.65cm}
  \label{fig1}
\end{figure}
Several studies~\cite{Hutcheon2000,Pulvermüller2013} have investigated the activation mechanisms of brain neurons and reveal that different tasks will activate different regions of the brain, as shown in Fig~\ref{fig1}. Motivated by such a phenomenon, we propose a novel parameter sharing technique called \textbf{Ada}tive \textbf{P}arameter \textbf{S}haring~(AdaPS). We first use VAE to extract the identity vectors of each agent, then cluster the agents into $K$ classes based on identity vectors, the cluster center will be input into a mapping network to generate an adaptive mask which corresponding to its own subnetwork. Agents in one cluster can share the experience to promote learning due to they have the same network structure. Since the AdaPS does not introduce additional training parameters during RL training, it ensures both scalability and high sample efficiency. The experimental results across multiple environments demonstrate that our method can ensure close to or even better performance while significantly reducing the number of parameters.


\section{Background}
\label{sec:background}

\subsection{Problem Formulation}
The multi-agent reinforcement learning problem in our paper can be modeled as a partially observed Markov game (POMG). A typically POMG can be defined by a tuple $U = \left< S,\{A^i\}_{i\in \mathcal{N}},\mathcal{N},\{O^i\}_{i\in\mathcal{N}},\mathcal{P}, \{R^i\}_{i\in\mathcal{N}}\right>$. Here, $\mathcal{N}$ represents the set of agents, $S$ is the state space of the environment. $\boldsymbol{A}=\prod_{i\in\mathcal{N}}A^i$ is the joint action space, and $\mathcal{O} = \prod_{i\in\mathcal{N}}{O^i}$ is the joint observation space. At each discrete time step $t$, each agent $i\in\mathcal{N} = \{1,...,N\}$ will select an action $a_i\in A_i$ according to its observation $o^i_t \in O$ which decided by current state $s_t$. The transition function $\mathcal{P}: S\times A\times S\rightarrow [0,1]$ will generate the next state $s_{t+1}$ given joint action $\boldsymbol{a}_t = (a^1_t,\cdots, a^N_t)$ and current state $s_t$. The reward function $R_{i}$ of each agent will give the agent its individual reward $r_t^i$ at time step $t$. The return $G_{i}=\sum^T_{t= 0}\gamma^tr^{t}_i$ is defined as the discounted sum of rewards, where $\gamma\in [0,1]$ is the discount factor. The goal of agent $i$ is to learn the policy $\pi_i(a_i|o_{i}):O^i \rightarrow P(A^i)$ to maximize its expected return. 

\subsection{Policy Gradient Methods}
The policy gradient (PG) method is a reinforcement learning technique that can directly learn a policy $\pi_\phi$ parameterized by $\phi$ for agents by optimizing its returns. The actor-critic algorithm is a simple and typical PG method. In the actor-critic algorithms, the actor is represented by a policy $\pi_\phi$ that selects actions based on the current observation, and the critic uses a value function $V(s;\omega)$ to estimate the expected rewards of the current state. Under the POMG setting, the policy loss function of an actor-critic algorithm can be defined as:
\begin{equation}
\begin{split}
    \mathcal{L}(\phi_i) = -log\ \pi(a^i_t|o^i_t;\phi_i)(r_t^i+\gamma V(o_{t+1}^i;\omega_i) - V(o_t^i;\omega_i)),
\end{split}
\end{equation}
and the value loss function of is:
\begin{equation}
\begin{split}
    \mathcal{L}(\omega_i) = ||V(o^i_t;\omega_i)-y_i||^2,
\end{split}
\end{equation}
where $y = r^i_t + \gamma V(o^i_{t+1};\omega_i)$. In our paper, we use A2C~\cite{mnih2016asynchronous} as our base RL method.

\subsection{Selective Parameter Sharing}
\label{seps}
Selective Parameter Sharing (SePS)~\cite{seps} uses Variational Autoencoders (VAEs) to encode the identity of agents, and the obtained identity vectors are used for clustering and learning separate networks for each category. The entire encoding process consists of two networks: an encoder $f_e$ and a decoder $f_d$, with parameters $\theta$ and $\beta$ respectively. The input to the encoder only includes the agent's ID, and the output is a $m$-dimension Gaussian distribution. Sampling from this distribution yields the agent's feature vector $z$. The decoder takes the agent's observations, actions and the agent's feature vector $z$ as input, and predicts the next observation and reward. Since the encoder's input does not include observation and action information, the vector $z$ only encodes the agent's intrinsic characteristics, such as the reward function $R$ and the transition model $P$. The agent's identity and observation transition distribution can be projected onto the latent space $Z$ based on the posteriors $q(z|i)$ and $p(z|tr=(o_{t+1},o_t,r_t,a_a))$. To find the posterior $q(z|i)$, the VAEs are employed to optimize the objective $D_{KL}(q_{\theta}(z|i)||p(z|tr))$, leading to a lower bound on the log-evidence of the transition $log\ p(tr)$ as follows:
\begin{equation}
\begin{split}
    log\ p(tr)\geq \mathbb{E}_{z\sim q_{\theta(z|i)}}[log\ p_\beta(tr|z)] - D_{KL}(q_\theta(z|i)||p(z)).
\end{split}
\label{dkl}
\end{equation}
Here, the first term, namely the reconstruction error, can be further decomposed as:
\begin{equation}
\begin{split}
    log&\ p_{\beta}(tr|z)  = log\ p_{\beta}(r_t,o_{t+1}|a_t,o_t,z)p(a_t,o_t|z) \\
     &= log\ p_{\beta}(r_t|o_{t+1},a_t,o_t,z) + log\ p_{\beta}(o_{t+1}|a_t,o_t,z) + c,
\end{split}
\end{equation}
where $q_{\theta}$ and $p_{\beta}$ function as $f_e$ and $f_d$. To minimize the KL divergence, it is sufficient to maximize the lower bound of evidence. 

\section{Related Work}
\label{sec:related}

\subsection{Parameter Sharing}
One important technique used to address scalability issues in different fields is parameter sharing~\cite{gupta2017cooperative,chu2017parameter,yao2023ndc,yao2023improving}. Terry et al.~\cite{2020Revisiting} have theoretically analyzed the importance of parameter sharing. Parameter sharing ensures that the model size remains unchanged as the number of agents changes by sharing the same parameterized function among multiple agents. However, the effectiveness of simple parameter sharing usually relies on agents having similar observations and tasks. 
Recent MARL algorithms~\cite{qmix,zhang2023inducing,li2023explicit,xu2022mingling} also added a one-hot index of agents as model inputs to differentiate different agents. Nevertheless, some recent research~\cite{seps,kim2023parameter} has shown that only adding an index has limited expressiveness. To address this issue, SePS~\cite{seps} trains independent policy for each type of agent, but the model size still increases with the number of types. SNP-PS~\cite{kim2023parameter} randomly generates a pruned network for each agent and controls the weight sharing ratio through the pruning rate. However, SNP-PS overlooks the correlation among different agents due to the random generation of subnetworks. Our methods consider both agents' correlation and the parameter size.
 
\subsection{Neural Network Masking Mechanism}
Neural network masking techniques have been widely applied in various fields, with two classic approach including the Dropout~\cite{srivastava2014dropout} and the neural network pruning method~\cite{wang2021recent}. In the Dropout method, during the training process, a random mask is used to drop hidden neurons and their connections to prevent overfitting. Dropout's effectiveness can be attributed to the generation of diverse subnetworks through different masks. Consequently, Dropout can be regarded as an ensemble of multiple models, enhancing the overall generalization of the model. In our paper, the masking mechanism is employed to generate subnetworks for each cluster's agents, which improves the diversity ability among agents without increasing the model size. 

\vspace{-0.2cm}
\section{Methods}
\label{sec:method}
In this chapter, we first briefly introduce the overall framework of AdaPS, and then provide a detailed introduction to the different important components of AdaPS.

\subsection{The AdaPS Framework}
\begin{figure}[t]
  \centering
  \centerline{\includegraphics[width=8.8cm]{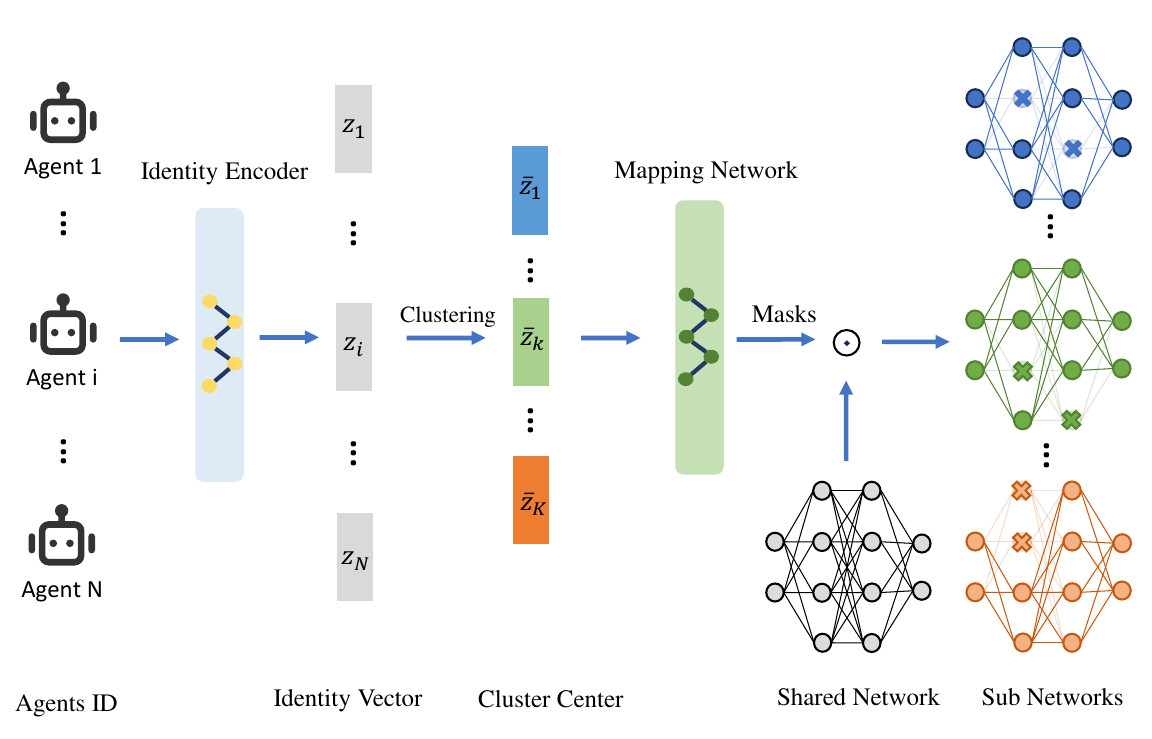}}
  \vspace{-0.2cm}
  \caption{The AdaPS framework.}
  \vspace{-0.2cm}
  \label{framework}
\end{figure}

We first train the VAE with the trajectories collected by the initial shared policy, so that it can encode the agents' identity information such as its transition model $P_i$ and reward function $R_i$. Then, we use K-means algorithm on the identity vector (sampled from output of encoder) to divide the agents into $K$ ($K<N$) clusters. The cluster center will feed into a mapping network to obtain its masks. The mask will be stored, so the mapping process only needs to be performed once at the beginning. During the RL training process, each agent will use its cluster's mask to obtain its own subnetwork. Therefore, we can construct the same network for similar agents without increasing the model size. The overall algorithm flow of AdaPS is illustrated in the Fig~\ref{framework}.

\subsection{Identity Endocing Process}
Learning from SePS~\cite{seps}, we employ a VAE to encode the identity of agents. The structure and learning objective of VAE can be found in Sec.~\ref{seps}. The amount of data required for this pre-training process is much smaller than the reinforcement learning process. We can input the index of agent $i$ into the encoder and output the identity vector $z_i$ by sampling. By further using clustering algorithms, agent with similar identity are classified into the same cluster. Next, we will describe the mapping function that can generate subnetworks for each cluster.

\begin{figure*}[ht]
\setlength{\belowcaptionskip}{-5cm} 
\subfigure[BPS (10-10-10)]{
\begin{minipage}{0.32\linewidth}
\centerline{\includegraphics[width=1.1\textwidth]{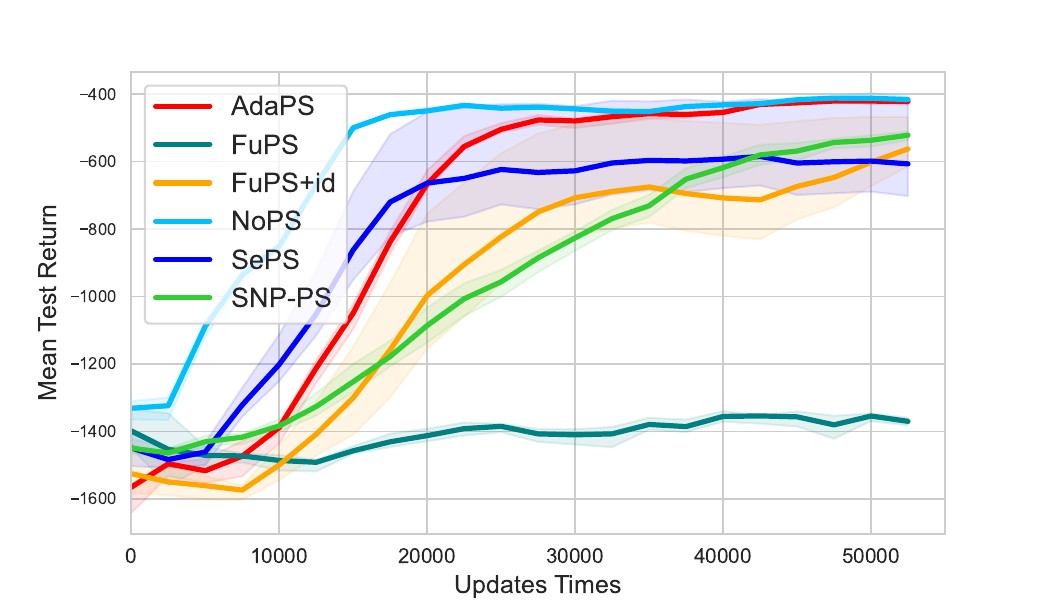}}
\label{fig:bps}
\end{minipage}}
\subfigure[LBF (3-3-3)]{
\begin{minipage}{0.32\linewidth}
\centerline{\includegraphics[width=1.1\textwidth]{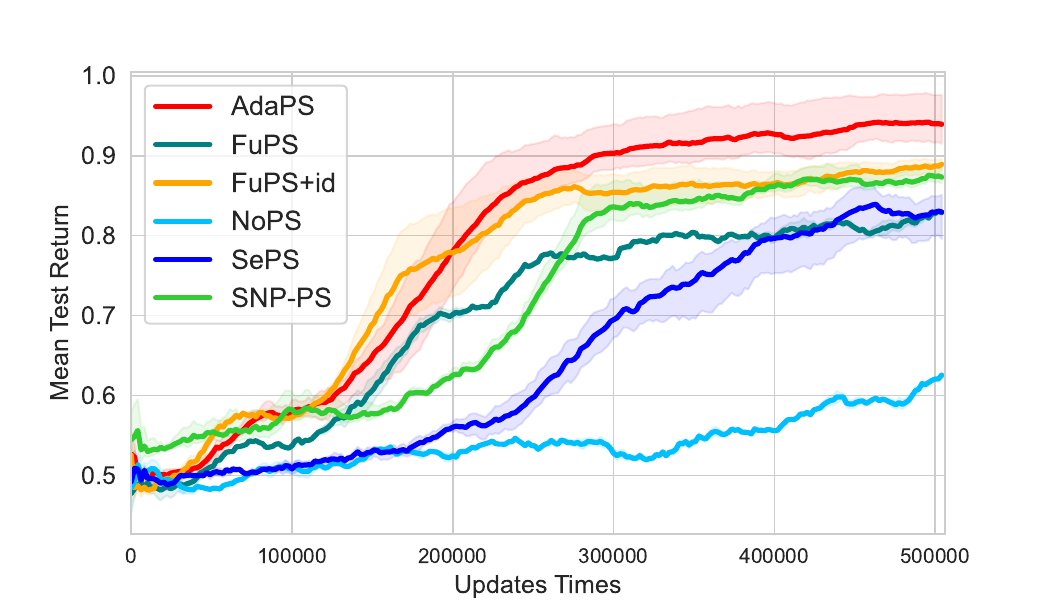}}
\label{fig:lbf}
\end{minipage}}
\subfigure[C-RWARE (2-2)]{
\begin{minipage}{0.32\linewidth}
\centerline{\includegraphics[width=1.1\textwidth]{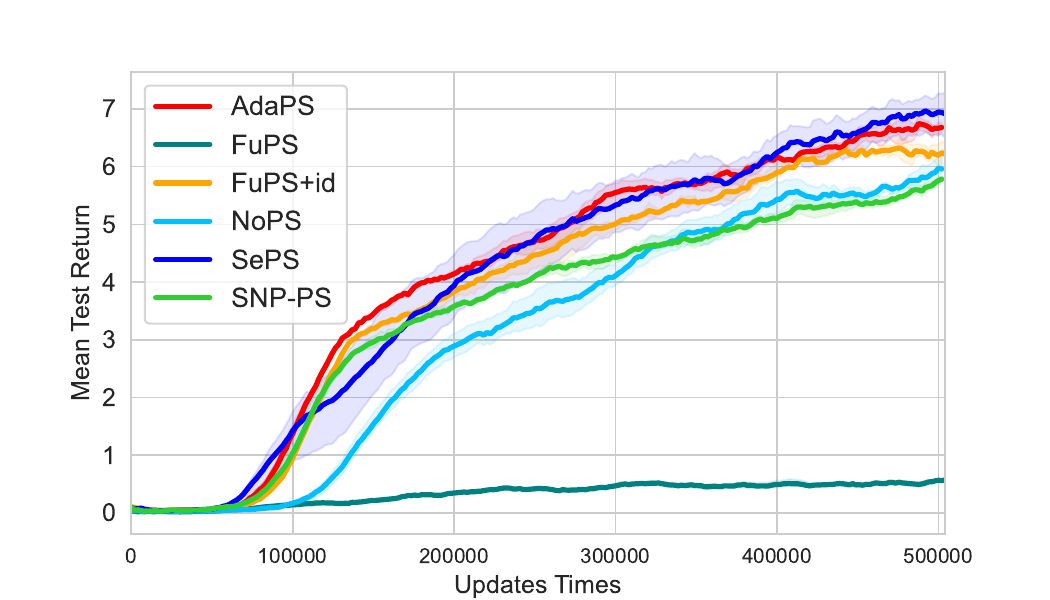}}
\label{fig:crware}
\end{minipage}}
\vspace{-0.2cm}
\caption{The mean test return on three environments. The numbers in bracket represent the number of agents in each type.}
\vspace{-0.2cm}
\label{fig:magent_curve}
\end{figure*}

\subsection{Network Mask Mapping}
We designed a mapping network which denoted as $f_m$ to map the average identity vector $\bar{z}$ (cluster center) of each cluster to different masks. The mapping network receives the vector of the cluster center and outputs the probability of activating each neuron in the hidden layer. The output dimension of the mapping network is equal to the number of neurons in the hidden layer. We choose a randomly initialized and fixed network as the mapping network~\cite{rnd}, since it is capable of generating uniform mappings without the need for training, thus avoiding any additional training cost. To relate the output of the mapping network to the activation probability of each neuron, we use sigmoid as the activation function of the mapping network thereby limiting the output of the network between 0 and 1. The masking generating process for cluster $k$ can be formalized as: 
 
\begin{equation}
    M_k = d(\text{Sigmoid}(f_m(\bar{z}_k))).
\end{equation}
The function $d$ can convert the activation probability to a binary mask:
\begin{align*}
\begin{split}
d(x)= \left \{
\begin{array}{ll}
1,                    & x > \lambda\\
0,                    & x \le \lambda
\end{array}
\right.
\end{split}
\end{align*}
and $\lambda$ is the drop threshold, the larger the $\lambda$, the more nodes in each subnetwork will be masked (or deactivated). In our experimental section, we simply set $\lambda=0.2$ for all scenarios. Assuming a shared policy network denoted as $f(\cdot;\phi)$, we can obtain the policy for cluster $k$ by using $f(\cdot ;\phi\odot M_k)$, where $\odot$ represents element-wise product. Through this masking process, agents in one cluster can get the same structures, enabling the sharing of experiences to facilitate learning. On the other hand, agents in different cluster get diverse strategies through different subnetwork structures.

\section{Experiments}
\label{sec:exp}
\subsection{Environment Description}
We use three multi-agent environments to evaluate the algorithms' performance. Each experiment is conducted with four random seeds, the shaded area is enclosed by the min and max value of all random seeds, and the solid line is the mean value of all seeds. We describe all environments below:

\textbf{Blind-particle Spread} 
The Blind-particle spread (BPS) is an environment modified based on the Multi-agent Particle Environment~\cite{lowe2017multi}. BPS consists of multiple agents and landmarks, each agent and landmark belongs to one color. The agents are not aware of their own color or the colors of other agents but need to navigate toward the correct landmark. 

\textbf{Coloured Multi-Robot Warehouse} 
The Coloured MultiRobot Warehouse (C-RWARE)~\cite{seps} includes multiple robots with distinct functionalities. The agents are tasked with delivering specific shelves, which are denoted by different colors, and have different action spaces. Agents receive a reward of 1.0 only when they successfully reach the goal with the requested shelf of the correct color, resulting in sparse rewards.

\textbf{Level-based Foraging} 
The Level-based Foraging (LBF) is a multi-agent environment~\cite{albrecht2015game} where agents forage for randomly scattered food on a grid. Each agent and food item is assigned a level at the start of each episode. LBF is partially observable, allowing agents to see the positions of other agents, food and its levels, but not the levels of other agents.

\vspace{-0.2cm}
\subsection{Configuration and Baseline}
We compare Adapts with various parameter sharing methods including 1) No Parameter Sharing (NoPS) - each agent has its own parameter without any overlap. 2) Full Parameter Sharing (FuPS) - full parameter sharing in which all agents share the same parameters without agent indication. 3) Full Parameter Sharing with index (FuPS+id) - full parameter sharing with one-hot encoding agent indication. 4) Selective Parameter Sharing (SePS) - the agents will be grouped by clustering algorithm and each group share one parameter network. 5) SNP-PS - generate a random mask for each agent. All baselines are implemented by A2C with n-step rewards.

\vspace{-0.2cm}
\subsection{Results on Different Environment}
The experiments in multiple scenarios are illustrated in the figure. In the BPS scenario (Fig.~\ref{fig:bps}), due to its simplicity, the NoPS algorithm achieve the best performance, while AdaPS show a similar performance to NoPS. The FuPS perform poorly because it cannot differentiate between different agents. In the LBF scenario (Fig.~\ref{fig:lbf}), where tasks among different agents are more similar, experience sharing among similar agents becomes crucial. AdaPS, capable of creating different substructures for each cluster while retaining the advantage of experience sharing, achieve optimal results. FuPS and FuPS+id perform slightly worse than AdaPS. SePS clustering agents into $K$ classes result in only $\frac{1}{K}$ samples in per class, leading to inefficient learning. NoPS, lacking any experience sharing among agents, perform poorly. In the C-RWARE scenario (Fig.~\ref{fig:crware}), where different agent categories have completely different tasks, FuPS exhibit lower performance, while other algorithms show similar performance.

\begin{figure}[t]
  \centering
  \centerline{\includegraphics[width=7cm]{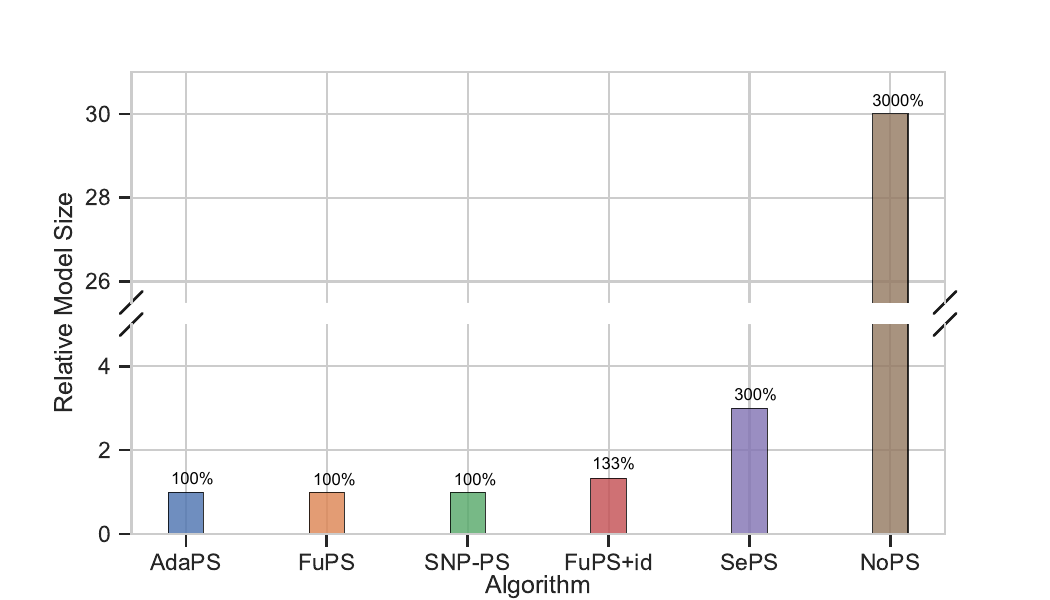}}
  \vspace{-0.4cm}
  \caption{Relative model size of different algorithms.}
  \vspace{-0.4cm}
  \label{fig:bar}
\end{figure}

\vspace{-0.2cm}
\subsection{More Analysis}
To further investigate the training costs of different models, we visualize the model sizes of each algorithm in the BPS scenario. There are three types of colored agents in the BPS scenario, with 10 agents per type. As shown in Fig.~\ref{fig:bar} , NoPS requires training independent models for each agent, resulting in large amounts of training costs (with 30 times the model size). The SePS also needs to train a separate model for each type of agent, causes 3 times the size of parameters. The AdaPS method has an identical training parameter quantity to that of FuPS since they both require only one common model to be trained. Among algorithms with similar parameter scales, AdaPS exhibit the best performance.

\vspace{-0.1cm}
\section{Conclusions}
In this paper, we propose a novel parameter sharing technique called AdaPS. The AdaPS method first uses VAEs to obtain identity vectors for each agent and partitions agents into clusters. Then we feed the cluster centers into a mapping function to generate different masks. By utilizing masks, AdaPS can construct diverse subnetwork for each cluster. Experimental results demonstrate the AdaPS method achieves higher sample efficiency and can generate policies with sufficient diversity without introducing additional training parameters.

\section{Acknowledgments}
This project was supported by Strategic Priority Research Program of the Chinese Academy of Sciences, Grant No. XDA27050100.

\small
\bibliographystyle{IEEEbib}
\bibliography{main}

\end{document}